\documentclass{article}

\usepackage{microtype}
\usepackage{graphicx}
\usepackage{subcaption}
\usepackage{booktabs} 

\usepackage{hyperref}

\usepackage[preprint]{icml2026}

\usepackage{amsmath}
\usepackage{amssymb}
\usepackage{mathtools}
\usepackage{amsthm}
\usepackage{algorithm}
\usepackage{algorithmic}
\usepackage{multirow}
\usepackage{booktabs}
\usepackage{adjustbox}
\usepackage[table]{xcolor}

\usepackage[capitalize,noabbrev]{cleveref}

\theoremstyle{plain}

\theoremstyle{definition}

\theoremstyle{remark}

\usepackage[textsize=tiny]{todonotes}

\icmltitlerunning{No Need for Learning to Defer?}
\newcommand{\ie}{\textit{i.e.}}
\newcommand{\eg}{\textit{e.g.}}
\definecolor{lightblue}{RGB}{173,216,230}

\usepackage{dsfont}
\begin{document}

\twocolumn[
  \icmltitle{No Need for Learning to Defer? A Training Free Deferral Framework to Multiple Experts through Conformal Prediction}



  \icmlsetsymbol{equal}{*}

  \begin{icmlauthorlist}
    \icmlauthor{Tim Bary}{ucl,saf}
    \icmlauthor{Beno\^it Macq}{ucl}
    \icmlauthor{Louis Petit}{saf}

  \end{icmlauthorlist}

  \icmlaffiliation{ucl}{ICTEAM, UCLouvain, Louvain-la-Neuve, Belgium}
  \icmlaffiliation{saf}{SAFiR Lab, University of Sherbrooke, Sherbrooke, Canada}

  \icmlcorrespondingauthor{Tim Bary}{tim.bary@uclouvain.be}

  \icmlkeywords{Learning to Defer, Conformal Prediction, Human-AI Teaming, Hybrid Intelligence}

  \vskip 0.3in
]



\printAffiliationsAndNotice{}  

\begin{abstract}
AI systems often struggle to provide reliable predictions across all inputs, motivating hybrid human-AI decision-making. Existing Learning to Defer (L2D) approaches address this by training models to selectively defer to human experts. However, these approaches require extensive training data annotated by all experts and are sensitive to changes in expert composition, necessitating costly retraining. We propose a training-free, model- and expert-agnostic framework for expert deferral based on conformal prediction. Our method leverages prediction sets from a conformal predictor to quantify label-specific uncertainty and selects the most suitable expert using a \textit{segregativity} criterion, which measures how well an expert discriminates among plausible labels. Experiments across three  models on CIFAR10-H and HAM10000 demonstrate that our method can reduce the number of training labels per expert by up to 91.3\% while maintaining predictive accuracy in low-data regimes. Being training-free, it also reduces training time by two orders of magnitude, offering a scalable, alternative to L2D for real-world human-AI collaboration.
\vspace{1em}

\end{abstract}

\section{Introduction}
Artificial Intelligence (AI) systems still require human oversight in high-stakes or ambiguous settings \cite{burell2016machine,dembrower2025human}. To address this limitation, Hybrid Intelligence (HI) frameworks, where AI collaborates with humans to achieve better joint performance than either could alone \cite{dellermann2019hybrid}, are increasingly being explored \cite{bhatt2025learning, boyd2023value,donahue2024two,radeta2024man}. One motivation for such systems is the observation that humans and AI tend to make uncorrelated errors, allowing humans to complement AI in areas of uncertainty \cite{liu2025human, hemmer2024complementarity}.

A major line of work in this space is \textit{Learning to Defer} (L2D) \cite{madras2018predict}, which trains a model to choose between making a prediction and deferring to a human. Although L2D methods have demonstrated strong performance, they require extensive labeled data, including expert annotations, and are tightly coupled to the experts they are trained with. These constraints limit their adaptability in practical deployments, where expert teams and their performances evolve over time, and where costly expert annotations are better allocated to new cases rather than model retraining.

An alternative path is to base collaboration on explicit uncertainty estimates rather than learned deferral policies, as a core requirement for effective deferral is the ability to recognize when model predictions are unreliable \cite{li2023modeling,ruggieri2025things}. Conformal Prediction~(CP) provides a principled, model-agnostic way to quantify uncertainty by returning a set of plausible labels for each input. These sets are guaranteed to contain the true label with a user-specified probability under minimal assumptions, making CP well-suited to support human-AI decision-making.

In this work, we present a training-free, model- and expert-agnostic framework for deferring to human experts. Our method uses the prediction set generated by a conformal predictor to identify candidate labels and selects the most suitable expert based on a metric we introduce, called \textit{segregativity}. This score quantifies an expert’s ability to distinguish between plausible labels identified by an off-the-shelf model, which enables targeted and efficient querying.

We evaluate our method on two multi-expert datasets: CIFAR10-H \cite{peterson2019human} and HAM10000 \cite{tschandl2018ham10000}. Using only a fraction of the expert-labeled training data, our framework achieves competitive accuracy against common L2D baselines on CIFAR10 and outperforms all baselines on two models on HAM10000. In low-data regimes, our method requires only 8.7\% of the expert labels needed by L2D approaches to surpass their accuracy.

\paragraph{Contributions}
\begin{itemize}
    \item We propose a training-free, model- and expert-agnostic deferral framework for human-AI collaboration based on conformal prediction.
    \item We introduce \textit{segregativity}, a label set-specific metric that selects the most suitable expert based on their ability to resolve ambiguity within a conformal prediction set.
    \item We validate our approach on two datasets, three model architectures, and five conformal scoring functions, achieving accuracy comparable to L2D baselines while using up to $91.3\%$ fewer expert-labeled training examples and reducing the training time by two orders of magnitude.
    \item We provide an extensive empirical analysis on CIFAR10, identifying the data regimes in which traditional L2D approaches become preferable to our method.
\end{itemize}

Code will be made publicly available.
\section{Related Works}
\subsection{The Learning to Defer Framework}\label{sec:L2D}
This section outlines the progression of L2D research and highlights the main theoretical and practical advancements.

\paragraph{Foundational Work and Single-Expert Settings} Initial formulations of L2D extended traditional rejection learning by incorporating human decision-making into the training objective. \cite{madras2018predict} introduced a framework that compared model and human confidence to guide deferral, enabling AI systems to delegate decisions when humans were likely to be more accurate. However, this early work was limited by the inconsistency of its loss function and the need for accurate estimates of human uncertainty.

\cite{mozannar2020consistent} addressed these issues by proposing a consistent surrogate loss for multiclass deferral. Subsequent work by \cite{verma2022calibrated} focused on calibration, proposing a one-vs-all (OvA) surrogate loss that provided both consistency and expert-calibrated deferral probabilities, critical for reliability in high-stakes settings. \cite{mozannar2023should} further refined the theory by introducing the RealizableSurrogate, addressing realizable H-consistency (\ie, the requirement that all models consistent with the training data make similar predictions across domains) under hypothesis constraints and proving the computational hardness of deferral learning in linear settings. \cite{wei2024exploiting} challenged the independence assumption between human and model predictions, introducing a dependent Bayes-optimal formulation and the Dependent Cross-Entropy loss to explicitly model human-AI interaction.

\paragraph{Reducing Expert Supervision and Annotation Cost} Recent work has focused on minimizing the reliance on full expert annotations. \cite{hemmer2023learning} introduced a three-stage semi-supervised method to approximate expert decisions and reduce annotation requirements. \cite{zhang2023learning} created LECOMH, a framework that can train exclusively on noisy expert labels, without requiring ground truth. \cite{nguyen2025probabilistic} proposed Probabilistic L2D, a framework that uses Expectation-Maximization to learn under missing annotations. \cite{strong2025expert} developed EA-L2D, a Bayesian approach that eliminates the need for a fully expert-labeled training set, enabling generalization to unseen experts using small context sets.

\paragraph{Workload Control} Managing expert workload is of primary importance to avoid mistakes due to fatigue and overworking. With Probabilistic L2D, \cite{nguyen2025probabilistic} integrated workload constraints into the EM optimization of their probabilistic model. \cite{alvescost} introduced DeCCaF, a cost-sensitive L2D framework that accounts for both expert capacity and asymmetric error costs, optimizing allocation under real-world resource constraints. Finally, \cite{ponomarev2024simple} uses an heuristic to impose constraints on the deferral model, limiting the proportion of data that can be deferred to humans.

\paragraph{Generalization to Unseen Experts} \cite{tailor2024learning} addressed the challenge of generalizing to unseen experts through L2D-Pop, a meta-learning strategy that infers deferral policies from small context sets drawn from a population of experts. Similarly, EA-L2D \cite{strong2025expert} formalizes an expert-agnostic deferral approach, constructing explicit, interpretable Bayesian representations of expert performance for out-of-distribution robustness.

\paragraph{Multi-Expert Extensions} Recognizing the prevalence of multi-expert settings in practice, \cite{keswani2021towards} and \cite{ijcai2022p344} extended L2D, which focussed on deferring to a single expert, to scenarios involving multiple human experts. In particular, \cite{ijcai2022p344} proposed a joint training framework to optimize overall team performance. However, this mixture-of-experts strategy was shown to be theoretically inconsistent. \cite{verma2023learning} addressed this by deriving consistent surrogate losses for multi-expert deferral, leveraging both softmax and OvA parameterizations. They also introduced conformal inference for principled expert ensembling and further analyzed calibration in the multi-expert case.

\paragraph{Two-Stage Learning for Pretrained Predictors} L2D-derived frameworks traditionally train the classification model and the deferrer conjointly. To enable practical deployment with fixed, pretrained models, \cite{mao2023two} proposed a two-stage learning paradigm. They first train a predictor using standard objectives (\eg, cross-entropy), and then train a deferral module with novel surrogate losses, which are shown to be H-consistent.

\subsection{Conformal Prediction}
Conformal Prediction (CP) is a model-agnostic framework for uncertainty quantification that constructs prediction sets rather than single-label outputs. These sets are guaranteed to contain the true label with user-specified marginal probability $1 - \alpha$, assuming exchangeable data. To do this, CP uses a separate calibration set with known labels to compute a conformal score for each sample, measuring how "atypical" the true label appears under the model's output. A quantile of these scores is then computed to define a threshold. At test time, the model forms a prediction set by including all labels whose scores fall below this threshold. The size of the prediction set adapts to the model’s confidence: inputs with high certainty yield small (often singleton) sets, while ambiguous inputs produce larger ones.

Different CP variants define different scoring rules. The Least Ambiguous Classifier (LAC) \cite{sadinle2019least} uses the inverse of the model’s confidence in the true label; it tends to produce small sets, but may return empty ones for ambiguous inputs. The Adaptive Prediction Sets (APS) method \cite{romano2020classification} uses the cumulative probability up to the true label in the sorted prediction vector. While APS often avoids empty sets in practice, randomized implementations can still result in them in edge cases. Regularized APS (RAPS) \cite{angelopoulos2020uncertainty} extends APS with penalties for large or low-ranked labels, balancing set size and coverage through two tunable parameters.

For a more detailed and accessible overview of conformal prediction in the context of deep learning, we refer readers to the tutorial by Angelopoulos and Bates \cite{angelopoulos2021gentle}.

\subsection{Conformal Prediction for Hybrid Intelligence}
A few works have explored the use of prediction sets generated via CP to support human experts in multiclass decision tasks. This approach aims to reduce expert misjudgment by constraining the human choices within the model's output conformal sets.

\cite{babbar2022utility} combined the L2D framework with CP to further improve the human performances and reduce the prediction set size for non-deferred samples. This work used a \textit{lenient} decision set, meaning that the expert may override it and pick a class outside of the set.

\cite{straitouri2023improving} pioneered \textit{strict} CP-based systems that require experts to choose only from the suggested label set, showing that this enforced structure can significantly improve expert accuracy and robustness, outperforming lenient designs. Building on this, their subsequent works have developed efficient bandit algorithms for optimizing prediction sets to maximize performances \cite{straitouri2024designing} and studied trade-offs between predictive performance and counterfactual harm to experts \cite{straitouri2024controlling}.

Other contributions in CP for human-AI teams include computational hardness results for optimal set construction \cite{de2024towards}, empirical validations of human performance gains \cite{cresswell2024conformal}, and theoretical critiques of CP’s interpretability and assumptions \cite{hullman2025conformal}. 

In summary, contributions in the field of CP for HI have focused on demonstrating the foundational benefits of enforced choice within sets, optimizing performance under imposed models, and rigorously investigating real-world human behavior. However, the entirety of the previously cited art solely focused on the relationship between a model and a single human expert. In this paper, we extend this relationship to a dynamic crowd with variable expertise. To the best of our knowledge, this is the first work to use CP for training-free, \emph{multi-expert} deferral.

\section{Proposed Deferral Framework}\label{sec:deferral}

We consider a setting where a predictive model $\phi$ and a pool of experts $\{1,\dots, K\}$ are available. For each expert $k$, we are given a set $\mathcal{Y}_k$ consisting of tuples $(\hat{y}_k,y)$, respectively the expert’s prediction and the true label on input data. This set allows estimation of each expert’s confusion patterns. We further assume access to a calibration set \( \mathcal{D}_{\text{cal}} \) of image-labels pairs. Note that the data sources sampled for calibration or expert evaluation do not need to match, as long as their distributions are exchangeable with the test distribution. Moreover, we do not necessarily assume equal sizes between $\mathcal{D}_{\text{cal}}$ and any of the $\mathcal{Y}_k$ datasets.

At test time, for each new data point $x$, our system must decide between:
\begin{enumerate}
    \item Accepting the model's prediction, if it is deemed sufficiently confident;
    \item Deferring to a human expert, selected from the pool based on their suitability for the case.
\end{enumerate}

To quantify the model uncertainty, we leverage $\mathcal{D}_{\text{cal}}$ to construct a conformal predictor around $\phi$. For the remainder of this article, we simply define a conformal predictor with miscoverage rate $\alpha$ as a function $\Gamma_\alpha(\cdot)$, which maps an input $x$ into a set  of plausible labels $\Gamma_\alpha(x)$ such that:
\begin{equation}
    \mathbb{P}(y\in \Gamma_\alpha(x)) \geq 1-\alpha
\end{equation}

If $|\Gamma_\alpha(x)|=1$, the model is considered sufficiently confident and we accept its prediction. Otherwise, we examine $\Gamma_\alpha(x)$ to evaluate to which expert $x$ should be deferred for annotation.

\subsection{Deferring to an Expert}\label{sec:experts}

Given $|\Gamma_\alpha(x)|>1$, we evaluate the suitability of each expert $k$ by restricting their assessment dataset $\mathcal{Y}_k$ to the labels deemed plausible by the model. Specifically, we define:
\begin{equation}
    \check{\mathcal{Y}}_k\coloneqq\left\{ (\hat{y}_k, y) \in\mathcal{Y}_k : \hat{y}_k\in \Gamma_\alpha(x) \land y\in \Gamma_\alpha(x)\right\},
\end{equation}
which contains only the expert predictions and ground-truth labels that lie within the conformal set. This restriction isolates the expert’s behavior on the subset of classes that remain ambiguous for the model and discards performance on irrelevant labels.

\begin{figure*}
    \centering
    \includegraphics[width=\linewidth]{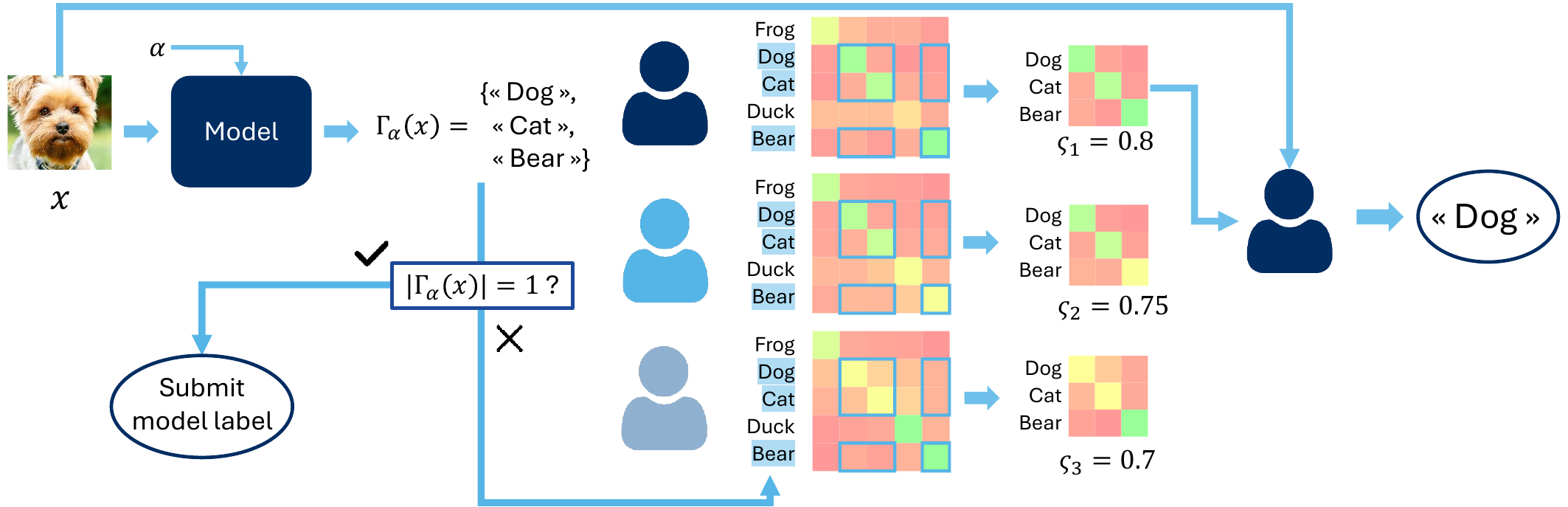}
    \caption{Our proposed deferral framework.  Given an input $x$, a conformal predictor based on a pre-trained model produces a prediction set $\Gamma_\alpha(x)$. If $|\Gamma_\alpha(x)|>1$, the decision is deferred to the expert with the highest segregativity $\varsigma_k$. An expert’s segregativity is defined as its accuracy on the sub-matrix of its confusion matrix restricted to the labels within $\Gamma_\alpha(x)$.}
    \label{fig:method}
\end{figure*}

We then quantify the ability of expert $k$ to resolve this ambiguity as its \textit{segregativity}, denoted $\varsigma_k$. If $|\check{\mathcal{Y}}_k| = 0$, we set $\varsigma_k = 0$. Otherwise, segregativity is defined as the empirical accuracy of expert $k$ over $\check{\mathcal{Y}}k$:
\begin{equation}\label{eq:segregativity}
    \varsigma_k (\Gamma_\alpha(x)) \coloneqq \frac{1}{|\check{\mathcal{Y}}_k|}\sum_{(\hat{y}_k, y)\in \check{\mathcal{Y}}_k} \mathds{1}_{\hat{y}_k = y},
\end{equation}
where $\mathds{1}$ is the indicator function. Intuitively, segregativity measures how well an expert discriminates among the labels contained in $\Gamma_\alpha(x)$, independently of its performance on other classes. Equivalently, $\varsigma_k$ can be computed by extracting the sub-matrix of the expert’s confusion matrix corresponding to the labels in $\Gamma_\alpha(x)$ and evaluating its accuracy (see Fig.~\ref{fig:method}).

The system selects the expert with the highest segregativity to provide the final prediction. In practice, expert assessment datasets are often small due to the cost of human annotation, resulting in coarse-grained segregativity estimates and frequent ties. These ties can be resolved according to system priorities, for instance by selecting the most accurate expert overall or the least costly one among those tied.

Under certain configurations, especially with high miscoverage rate $\alpha$, $\Gamma_\alpha(x)$ may be empty. When using the LAC conformal score, this typically occurs for highly ambiguous inputs, such as those near decision boundaries. In this case, we set $\check{\mathcal{Y}}_k = \mathcal{Y}_k$ for all experts, effectively defaulting to the expert with the highest overall accuracy. In contrast, empty prediction sets under APS or RAPS scoring usually indicate extreme model confidence. In such cases, the model’s prediction is accepted without deferral. 

\subsection{Motivations}\label{sec:motivations}

Similarly to previous works \cite{babbar2022utility, straitouri2023improving}, our framework builds on the use of CP sets to support selective deferral, leveraging their marginal coverage guarantees to identify inputs that can be handled autonomously by the model while maintaining high reliability. When the conformal prediction set $\Gamma_\alpha(x)$ is a singleton, the conformal predictor provides strong evidence of model reliability. Ambiguous inputs, characterized by larger prediction sets, are instead deferred to human experts.

When deferral is required, our method selects experts based on their ability to resolve ambiguity within the specific subset of labels identified by $\Gamma_\alpha(x)$. Privileging general accuracy may obscure such specialization, as a specialized expert may be highly effective at distinguishing among a small group of fine-grained or visually similar classes, yet have a limited accuracy due to errors on unrelated labels. By restricting the decision space to $\Gamma_\alpha(x)$, our framework enables delegation to such specialists when appropriate. Conversely, when the model is highly uncertain and $\Gamma_\alpha(x)$ is large, the framework naturally favors more generalist experts. As a result, the balance between specialist and generalist expertise emerges organically from the size of the conformal prediction set.

The miscoverage rate $\alpha$ plays a central role in controlling both system accuracy and expert workload. Larger values of $\alpha$ produce smaller prediction sets on average, allowing more inputs to be handled directly by the model but potentially at the cost of reduced accuracy\footnote{Note that for LAC conformal scores, an increase in $\alpha$ beyond a certain threshold instead leads to inputs being deferred to the experts, as more empty sets are generated.}. Conversely, smaller values of $\alpha$ increase the expected size of $\Gamma_\alpha(x)$, leading to more frequent deferral to experts while ensuring that model-only decisions are highly reliable. However, when $\alpha$ becomes too small, conformal sets may include most or all labels, reducing their discriminative value and reverting the system to selecting the most accurate expert. This behavior suggests the existence of an optimal miscoverage level $\alpha^* \in [0,1]$ that maximizes overall system accuracy, with values $\alpha > \alpha^*$ trading accuracy for reduced expert intervention.

We define the proposed framework as \emph{expert-agnostic}, as experts can be seamlessly added or removed by providing their assessment datasets $\mathcal{Y}_k$, without modifications of the system structure. It is also \emph{model-agnostic} and \emph{training-free}, as it relies solely on an off-the-shelf predictive model wrapped with a conformal predictor.

\section{Experimental Setup}\label{setup}

\paragraph{Datasets}

We evaluate our approach on two multi-annotated datasets used in recent L2D studies \cite{mozannar2023should, zhang2023learning, verma2023learning}.

CIFAR10-H \cite{peterson2019human} contains annotations from 2,571 human participants, each labeling 200 images from the CIFAR10 test set. This results in 47-63 annotations per image. The average annotator accuracy is $94.87 \pm 5.28\%$. To allow training of the baselines, we additionally generate expert annotations for the CIFAR10 training set by sampling from the empirical confusion distributions of all experts.

HAM10000 \cite{tschandl2018ham10000} consists of 10,000 dermatoscopic images spanning 7 skin lesion classes. As the dataset does not include human annotations, we generate synthetic experts following the protocol of \cite{verma2023learning}. This results in 9 experts with accuracies ranging from 14.29\% to 80\%.

\paragraph{Baselines}

We compare our method against four multi-expert L2D baselines. Single-expert baselines are omitted, as they are consistently outperformed in multi-expert settings. Included baselines are \cite{ijcai2022p344} and \cite{verma2023learning}, both of which rely on learned surrogate losses for expert deferral. For each baseline, we also include a pre-trained variant following the two-stage paradigm of \cite{mao2023two}, where the classifier is first trained using standard objectives and the deferral module is subsequently fine-tuned.

To contextualize the benefits of hybrid intelligence, we additionally report the accuracy of the model alone and the accuracy of the best expert selected in hindsight.

\paragraph{Predictive Models}

We use three standard architectures as predictive models $\phi$: ResNet-18, VGG-11-BN, and DenseNet-161. For HAM10000, we use the default PyTorch 2.5.1 implementations. For CIFAR10-H, we adopt the architectures and training procedures from \cite{phan2021cifar10} with default settings.

For HAM10000, models are trained on 60\% of the dataset, with the remaining 40\% split into 10\% validation, 10\% calibration, and 20\% test sets. For CIFAR10-H, models are trained on 90\% of the training data, with the remaining 10\% split into 8\% validation and 2\% calibration sets. All models are trained for 50 epochs across all experiments. Model-only accuracies are reported in Table~\ref{tab:benchmarks}.

\paragraph{Expert Supervision and Label Budgets}

L2D baselines require training and validation on datasets fully annotated by all experts, resulting in 8,000 and 50,000 labeled samples per expert for HAM10000 and CIFAR10-H, respectively.

In contrast, our framework only requires small expert assessment sets $\mathcal{Y}_k$ of 140 expert labels per expert for HAM10000 and 800 expert labels per expert for CIFAR10-H. We additionally use 700 samples from the calibration set of each dataset as a meta-validation set to select the optimal miscoverage rate $\alpha^*$. This brings the total number of expert labels per expert to 840 for HAM10000 and 1,500 for CIFAR10-H. Appendix~\ref{app:set_sizes} discusses the rationale behind these choices of set sizes.

\begin{figure*}[ht]
    \centering
    \begin{minipage}{0.25\linewidth}
        \includegraphics[width=\linewidth]{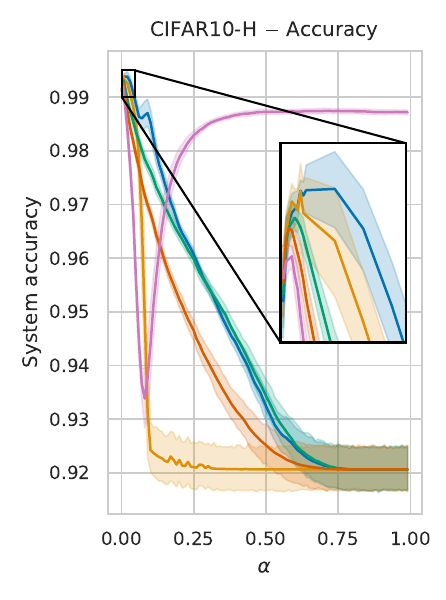}
    \end{minipage}%
    \begin{minipage}{0.25\linewidth}
        \includegraphics[width=\linewidth]{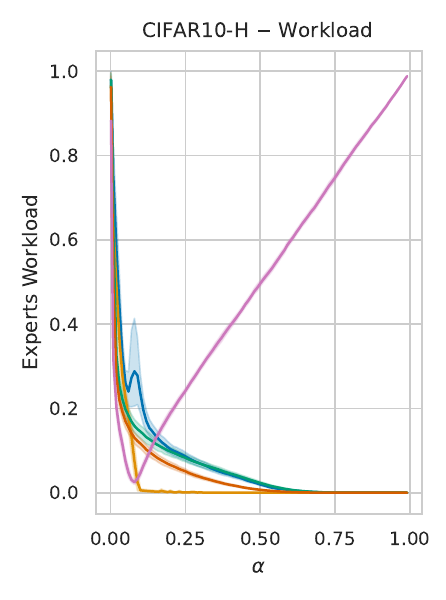}
    \end{minipage}%
    \begin{minipage}{0.25\linewidth}
        \includegraphics[width=\linewidth]{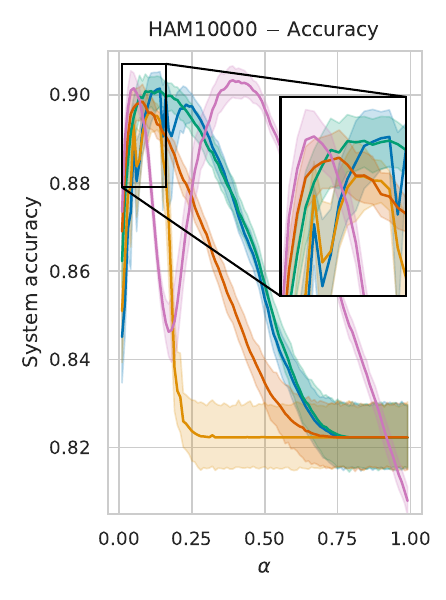}
    \end{minipage}%
    \begin{minipage}{0.25\linewidth}
        \includegraphics[width=\linewidth]{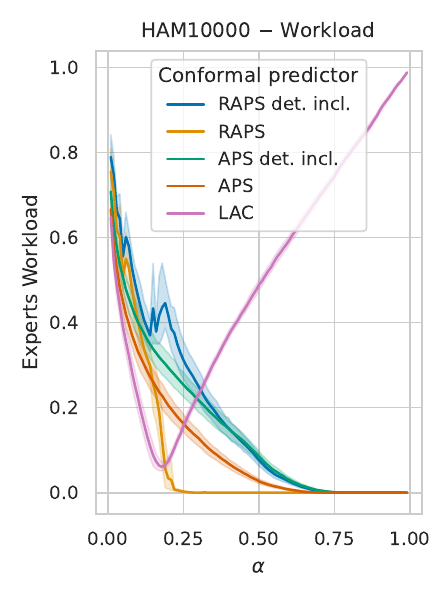}
    \end{minipage}
    \caption{Influence of the miscoverage rate $\alpha$ on system accuracy and expert workload for our framework on CIFAR10-H and HAM10000, across the considered conformal scoring functions. Results are averaged over five random seeds and three model architectures; per-model trends are consistent. Shaded regions indicate 95\% confidence intervals.}
    \label{fig:workload}
\end{figure*}

\paragraph{Conformal Prediction}

We construct conformal predictors using the Least Ambiguous Classifier (LAC) \cite{sadinle2019least}, Adaptive Prediction Sets (APS) \cite{romano2020classification}, and Regularized APS (RAPS) \cite{angelopoulos2020uncertainty}. For APS and RAPS, we consider both deterministic inclusion of the threshold-crossing label, preventing empty prediction sets at the cost of larger sets, and randomized inclusion, which preserves coverage guarantees but allows empty sets.

For both datasets, the calibration set $\mathcal{D}_{\text{cal}}$ consists of 1,000 stratified samples (10\% of HAM10000 and 2\% of CIFAR10-H). RAPS hyper-parameters are tuned on a held-out subset of $\mathcal{D}_{\text{cal}}$ following \cite{angelopoulos2020uncertainty}.

\section{Results and Discussion}
\subsection{Empirical Behavior of the Framework}

Figure~\ref{fig:workload} shows the evolution of system accuracy and expert workload as a function of the miscoverage rate $\alpha$ across the considered conformal predictors. The observed behavior closely matches the analysis of Section~\ref{sec:motivations}.

For APS and RAPS, decreasing $\alpha$ increases the average size of the conformal prediction sets, resulting in higher deferral rates and system accuracy approaching that of the best expert as $\alpha\to0$. Conversely, as $\alpha \to 1$, prediction sets and expert workload shrink and system accuracy converges to that of the model alone. Importantly, for a range of intermediate values of $\alpha$, the system consistently achieves accuracies exceeding both the model-only and best-expert performances, which demonstrates the benefit of adaptive human-AI collaboration.

The LAC predictor exhibits a distinct regime to APS and RAPS at larger values of $\alpha$. As $\alpha$ increases, LAC increasingly returns empty prediction sets, which, under our decision rule, trigger deferral to the most accurate expert. This behavior explains the observed convergence toward high expert workload and expert-level accuracy anticipated in Section~\ref{sec:experts}.

These results overall confirm that the miscoverage rate $\alpha$ provides an interpretable control over the accuracy-workload trade-off, while the qualitative trends remain consistent across datasets.

\subsection{Benchmarking Against L2D Baselines}

All benchmarking experiments are repeated across five random seeds, with identical data splits and expert annotations across methods for each seed. Our framework and the pre-trained L2D baselines use the same off-the-shelf models described in Section~\ref{setup}, while standard L2D baselines train models from random initialization. For our framework, the conformal score and $\alpha$ are selected using the meta-validation set described in Section~\ref{setup}. We perform a grid search over $\alpha \in [0.01, 0.99]$ with step size $0.01$, with additional refinement in $[0.001, 0.009]$ for CIFAR10-H, where the optimal region is concentrated (see Fig.~\ref{fig:workload}). Table~\ref{tab:benchmarks} reports final test accuracies and expert workloads.

On CIFAR10-H, our method outperforms the Hemmer baselines and remains within $0.29 \pm 0.16\%$ of the best-performing method (\ie Verma with pre-training) on average. This competitive performance is achieved using only $3\%$ of the expert labels required by learned L2D methods and reduces the time to system deployment by factors ranging from $144$ to $196$. Moreover, our framework relied predominantly on model predictions, yielding substantially lower expert workload ($0.49$–$0.64$) compared to L2D baselines, which systematically deferred inputs to human experts.

On HAM10000, despite using only $10.5\%$ of the expert labels and less than $0.5\%$ of the training time of the strongest baseline, our framework outperforms all L2D methods on ResNet-18 and DenseNet-161, and ranks second on VGG-11-BN. The higher expert workload observed relative to the pre-trained Verma baseline can be attributed to differences in underlying classifier quality. As this baseline fine-tunes the classifier for an additional 50 epochs, the standalone model accuracy improves by $3.76$–$4.93\%$ compared to the model used by our framework, which reduces the need for deferral.

\begin{table*}[t]
\caption{Comparison of the proposed framework with multi-expert L2D baselines on CIFAR10-H and HAM10000. Reported metrics include system accuracy, expert workload (fraction of test samples deferred), expert supervision cost (number of training labels per expert), and model training or calibration time. For each model and dataset, the best accuracy is shown in \textbf{bold}, and the second-best is \underline{underlined}. Standard deviations are reported after the $\pm$ symbol.}

\label{tab:benchmarks}
\aboverulesep = 0.1mm 
\belowrulesep = 0.1mm 

\begin{adjustbox}{width=\textwidth}
\begin{tabular}{l|cccc|cccc}
\toprule
\multirow{3}{*}{Strategy} 
& \multicolumn{4}{c|}{CIFAR10-H} 
& \multicolumn{4}{c}{HAM10000} \\
&Accuracy [\%] & Workload ($\downarrow$) & Training size ($\downarrow$) & Training time [s] ($\downarrow$)
& Accuracy [\%] & Workload ($\downarrow$) & Training size ($\downarrow$) & Training time [s] ($\downarrow$)\\


\midrule
\multicolumn{9}{c}{\textbf{(a) Results with ResNet-18}} \\
\midrule
Model alone & 91.68 $\pm$ 0.20 & 0.00 $\pm$ 0.00 & -& - & 81.15 $\pm$ 1.05 & 0.00 $\pm$ 0.00 & - & -\\
Best expert & 98.95 $\pm$  0.06 & 1.00 $\pm$ 0.00 & - & - & 80.58 $\pm$ 0.59 & 1.00 $\pm$ 0.00 & - & - \\
Hemmer & 96.31 $\pm$ 1.82 & 1.00 $\pm$ 0.00 & 50,000 & 893.24 $\pm$ 2.12&85.66 $\pm$ 0.80 & 0.64 $\pm$ 0.01& 8,000 & 644.08 $\pm$ 56.3 \\
Hemmer (pre-trained) & 98.68 $\pm$ 0.18 & 1.00 $\pm$ 0.00 & 50,000 & 892.17 $\pm$ 3.91& 85.39 $\pm$ 1.18& 0.26 $\pm$ 0.04 & 8,000 & 590.22 $\pm$ 4.35 \\
Verma & \underline{99.52} $\pm$ 0.04 & 1.00 $\pm$ 0.00 & 50,000 & 1755.11 $\pm$ 7.88 &84.48 $\pm$ 0.56 &0.25 $\pm$ 0.02 & 8,000 & 795.32 $\pm$ 3.12 \\
Verma (pre-trained) & \textbf{99.62} $\pm$ 0.04 & 1.00 $\pm$ 0.00 & 50,000 & 1728.59 $\pm$ 4.21& \underline{89.15} $\pm$ 0.30& 0.15 $\pm$ 0.01 & 8,000 & 787.68 $\pm$ 4.91 \\
Segregativity (Ours) \cellcolor{lightblue}&99.22 $\pm$ 0.21\cellcolor{lightblue} &0.49 $\pm$ 0.13\cellcolor{lightblue} &1,500\cellcolor{lightblue} &\cellcolor{lightblue} 11.99 $\pm$ 0.89&\textbf{89.58} $\pm$ 0.87\cellcolor{lightblue} &0.39 $\pm$ 0.09\cellcolor{lightblue} &\cellcolor{lightblue}840 &\cellcolor{lightblue} 3.72 $\pm$ 0.64\\


\midrule
\multicolumn{9}{c}{\textbf{(b) Results with VGG-11-BN}} \\
\midrule
Model alone & 91.35 $\pm$ 0.11 & 0.00 $\pm$ 0.00 & - & - & 81.69 $\pm$ 0.62 & 0.00 $\pm$ 0.00 & - & -\\
Best expert & 98.95 $\pm$  0.06 & 1.00 $\pm$ 0.00 &- & - & 80.58 $\pm$ 0.59 & 1.00 $\pm$ 0.00 & - & - \\
Hemmer &97.29 $\pm$ 2.79& 1.00 $\pm$ 0.00& 50,000 & 898.85 $\pm$ 4.23& 86.28 $\pm$ 0.81&0.63 $\pm$ 0.01 & 8,000 & 649.98 $\pm$ 8.23 \\
Hemmer (pre-trained) & 97.94 $\pm$ 0.71 &1.00 $\pm$ 0.00 & 50,000 & 893.27 $\pm$ 3.27&87.24 $\pm$ 1.60 &0.28 $\pm$ 0.07 & 8,000 & 625.55 $\pm$ 1.87 \\
Verma & \textbf{99.63 } $\pm$  0.02 &1.00 $\pm$ 0.00 & 50,000 & 1767.69 $\pm$ 5.70 & 84.42 $\pm$ 0.72& 0.12 $\pm$ 0.03& 8,000 & 803.25 $\pm$ 3.81 \\
Verma (pre-trained) & \textbf{99.63} $\pm$ 0.01 &1.00 $\pm$ 0.00 & 50,000 & 1765.56 $\pm$ 3.77 &\textbf{90.16} $\pm$ 0.27 &0.12 $\pm$ 0.01 & 8,000 & 802.75 $\pm$ 4.04 \\
Segregativity (Ours) \cellcolor{lightblue}&\underline{99.47} $\pm$ 0.12\cellcolor{lightblue} &0.61 $\pm$ 0.08\cellcolor{lightblue} &1,500\cellcolor{lightblue} &\cellcolor{lightblue} 12.21 $\pm$ 1.04&\underline{90.02} $\pm$ 0.52\cellcolor{lightblue} &0.37 $\pm$ 0.06\cellcolor{lightblue} &\cellcolor{lightblue}840 &\cellcolor{lightblue} 3.98 $\pm$ 0.32\\


\midrule
\multicolumn{9}{c}{\textbf{(c) Results with DenseNet-161}} \\
\midrule
Model alone &93.15 $\pm$ 0.29 & 0.00 $\pm$ 0.00 & - & - & 83.85 $\pm$ 1.12 & 0.00 $\pm$ 0.00 & - & -\\
Best expert & 98.95 $\pm$  0.06 & 1.00 $\pm$ 0.00 & - & - & 80.58 $\pm$ 0.59 & 1.00 $\pm$ 0.00 & - & - \\
Hemmer & 96.31 $\pm$ 1.82 &1.00 $\pm$ 0.00 & 50,000 & 1160.00 $\pm$ 7.07 & 86.23 $\pm$ 0.23& 0.63 $\pm$ 0.01& 8,000 & 805.34 $\pm$ 10.94 \\
Hemmer (pre-trained) & 98.63 $\pm$ 0.08 &1.00 $\pm$ 0.00 & 50,000 & 1171.68 $\pm$ 1.86&84.73 $\pm$ 1.08 &0.16 $\pm$ 0.09 & 8,000 &  769.84 $\pm$ 2.32\\
Verma & \textbf{99.64} $\pm$ 0.04 &1.00 $\pm$ 0.00 & 50,000 &2522.06 $\pm$ 11.16 &86.46 $\pm$ 0.81 &0.22 $\pm$ 0.02 & 8,000 & 1308.29 $\pm$ 5.72\\
Verma (pre-trained) & \underline{99.61} $\pm$ 0.05 &1.00 $\pm$ 0.00 & 50,000 & 2491.03 $\pm$ 10.83 &\underline{90.21} $\pm$ 0.38 &0.07 $\pm$ 0.02 & 8,000 & 1288.48 $\pm$ 5.15 \\
Segregativity (Ours) \cellcolor{lightblue}& 99.29 $\pm$ 0.04\cellcolor{lightblue} &0.64 $\pm$ 0.18\cellcolor{lightblue} &1,500\cellcolor{lightblue} &\cellcolor{lightblue} 13.05 $\pm$ 0.96&\textbf{90.82} $\pm$ 0.39\cellcolor{lightblue} &0.32 $\pm$ 0.04\cellcolor{lightblue} &\cellcolor{lightblue}840 &\cellcolor{lightblue} 4.71 $\pm$ 0.51\\

\bottomrule

\end{tabular}
\end{adjustbox}

\end{table*}

\subsection{Performance of L2D in Low-Data Regimes}

While learned L2D methods can achieve marginally higher peak accuracy when abundant expert annotations are available, their performance degrades substantially as expert supervision becomes scarce. In many practical settings, requiring each expert to annotate tens of thousands of samples is unrealistic. We therefore analyze how L2D baselines perform as expert-labeled data is progressively reduced, and estimate the regime in which segregativity-based deferral becomes preferable.

We train and validate the Verma baseline (with and without pre-training) on subsets of CIFAR10-H ranging from 2,000 to 50,000 expert-labeled samples, preserving the training-validation split ratios of Section~\ref{setup}. Each configuration is evaluated across five seeds. Other baselines and HAM10000 are omitted, as their accuracy crossover points with segregativity-based deferral lie beyond the ranges shown in Table~\ref{tab:benchmarks}.

For each seed $s$ and dataset size $n$, we define $\Delta_{s,n}$ as the accuracy difference between the L2D baseline and its seed-matched segregativity-based counterpart. We model this difference using the following mixed-effects regression:
\begin{equation}
    \Delta_{s,n} = \beta_0 + \beta_1\log_2n + \beta_2(\log_2 n)^2+u_s+\epsilon_{s,n},
\end{equation}
where $u_s \sim \mathcal{N}(0, \sigma^2_{\text{seed}})$ captures seed-level variability and $\epsilon_{s,n}$ represents observation noise. The quadratic term accounts for saturation effects near the accuracy ceiling. Parameters are estimated via restricted maximum likelihood.

\begin{figure}[ht]
    \centering
    \includegraphics[width=\linewidth]{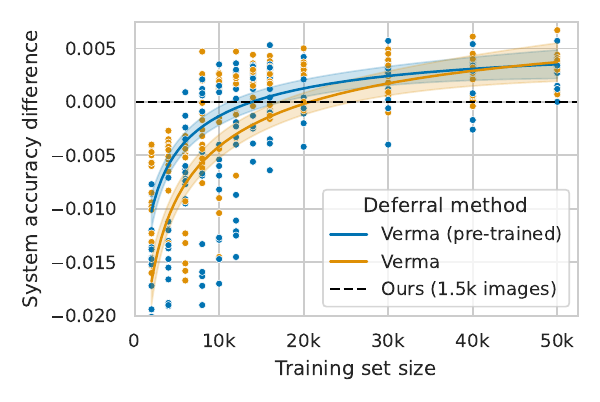}
    \caption{Accuracy difference between L2D baselines and segregativity-based deferral on CIFAR10-H as a function of the number of expert-labeled samples. Points show per-seed differences, while curves and shaded regions denote the fitted mixed-effects model and its 95\% confidence interval. The confidence bounds identify the dataset sizes at which L2D reliably matches or exceeds our framework’s accuracy using 1,500 expert labels.}
    \label{fig:degradation_info}
\end{figure}

Figure~\ref{fig:degradation_info} shows the observed accuracy differences alongside population-level predictions and 95\% confidence intervals. From these intervals, we estimate that the Verma L2D baseline requires between $11{,}407$ and $17{,}678$ expert-labeled samples with pre-training, and between $17{,}196$ and $25{,}879$ samples without pre-training, to reliably outperform our framework. Taking the lower bounds yields a reduction of up to 91.3\% in required expert annotations to match state-of-the-art performance in low-data regimes.

\subsection{Discussion}

We position our approach as a flexible and lightweight alternative to the Learning to Defer (L2D) framework, particularly suited to dynamic or resource-constrained deployment settings. A key limitation of L2D lies in its tight coupling to a fixed pool of experts: changes in expert availability, performance drift, or team composition typically require retraining the deferral model to preserve accuracy. This retraining incurs substantial computational cost and often demands extensive expert supervision, as many L2D methods rely on fully annotated training sets for calibration, despite recent efforts to relax this requirement in single-expert settings \cite{hemmer2023learning,alvescost,nguyen2025probabilistic}.

In contrast, our framework supports plug-and-play expert integration. Experts can be added or removed at test time using only their assessment datasets $\mathcal{Y}_k$, without modifying the underlying model or retraining any component. This design makes the framework both expert-agnostic and model-agnostic, enabling practitioners to freely update predictive models or incorporate new experts as they become available. In particular, our approach naturally accommodates large pretrained or foundation models whose retraining costs are prohibitive, opening the door to scalable hybrid intelligence systems.

This flexibility however comes with a trade-off. Our routing mechanism relies on label-conditional expert performance and implicitly assumes that an expert’s competence is homogeneous across instances of a given class. In contrast, learned L2D policies can capture finer-grained, instance-level interactions between inputs and expert performance, for example by exploiting auxiliary attributes such as dialect in hate speech detection \cite{ijcai2022p344,mozannar2020consistent}. Nevertheless, our empirical results demonstrate that label-conditional expertise is sufficient to achieve competitive accuracy across diverse datasets and architectures, suggesting that instance-level heterogeneity is not a limiting factor in many practical scenarios.

Overall, our results indicate that segregativity-based, training-free deferral provides a compelling alternative to learned deferral when adaptability, data efficiency, and rapid deployment are prioritized.

\section{Conclusion and Future Works}

We introduced a training-free, model- and expert-agnostic deferral framework that uses conformal prediction to quantify model uncertainty and a segregativity criterion to route ambiguous inputs to the most suitable expert. By relying solely on conformal sets and lightweight expert assessment datasets, our approach provides a flexible and data-efficient alternative to Learning to Defer (L2D) methods, particularly in settings where expert supervision and retraining are costly or impractical.

Across two multi-expert datasets and three model architectures, our framework achieves competitive or state-of-the-art system accuracy while requiring only a fraction of the expert labels and dramatically reducing time to deployment. In low-data regimes, we show that segregativity-based deferral can match or outperform learned L2D policies while reducing expert annotation requirements by up to 91.3\%.

Future work will explore extending this framework to scenarios with no prior expert assessments, leveraging patterns of expert disagreement and online feedback to infer expertise on the fly. Additional directions include adaptive expert modeling under non-stationary performance and deployment with large pretrained or foundation models, for which retraining-based deferral is infeasible.

\section*{Acknowledgments}
T.~Bary is funded by the Walloon region under grant No. 2010235 (ARIAC/TRAIL) and by Wallonia-Brussels International. Computational resources have been provided by the CÉCI, funded by the F.R.S.-FNRS under Grant No. 2.5020.11 and the Walloon Region. T.~Bary would like to thank Prof. Marcos Medeiros Raimundo for sparking his interest in conformal prediction.

\bibliography{main}
\bibliographystyle{icml2026}

\newpage
\appendix
\onecolumn

\section{Justification of Expert Assessment Sets and Meta-Validation Set Sizes}
\label{app:set_sizes}

Expert assessment sets $\mathcal{Y}_k$ are used to estimate expert confusion matrices, fundamental to the expert selection mechanism of our framework. Their design must balance statistical reliability and annotation cost. We build $\mathcal{Y}_k$ to satisfy two criteria.

First, class coverage must be uniform to avoid bias in the expert capacity to discriminate across a subset of classes, particularly for imbalanced datasets. To ensure this, assessment sets are sampled in \emph{shots}, where one shot consists of exactly one example per class. This corresponds to 10 samples per shot for CIFAR10-H and 7 samples per shot for HAM10000.

\begin{figure}[h]
    \centering
        \begin{minipage}{0.45\linewidth}
        \includegraphics[width=\linewidth]{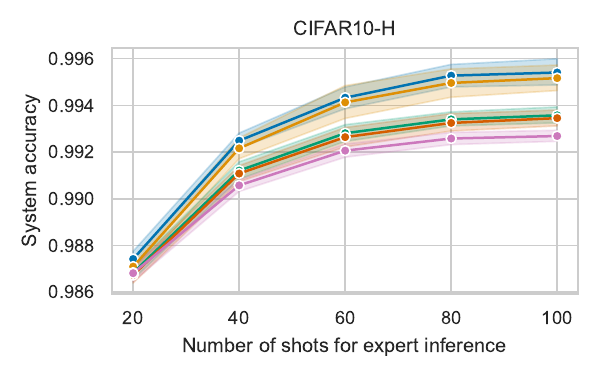}
    \end{minipage}%
        \begin{minipage}{0.45\linewidth}
        \includegraphics[width=\linewidth]{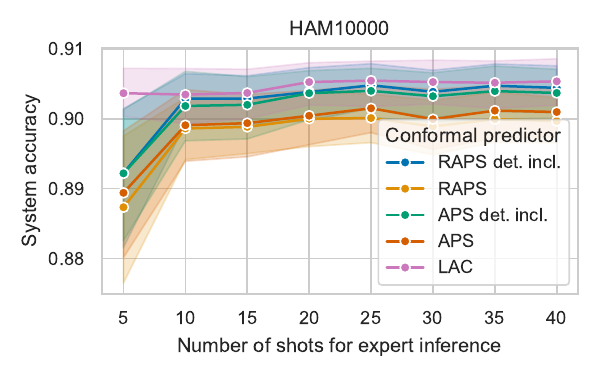}
    \end{minipage}%
    \caption{System accuracy of segregativity-based deferral at $\alpha^*$ as a function of expert assessment set size for CIFAR10-H and HAM10000 across conformal scoring functions. Results are averaged over five random seeds and three model architectures; per-model trends are consistent. Shaded regions denote 95\% confidence intervals.}
    \label{fig:n_shots}
\end{figure}

Second, the number of shots must be sufficient to reduce estimation variance while avoiding unnecessary annotation overhead, especially in low-data regimes. Figure~\ref{fig:n_shots} reports the system accuracy at the optimal miscoverage rate $\alpha^*$ as a function of the number of shots.
For CIFAR10-H, performance saturates at approximately 80 shots across all scoring methods, which we adopt in our experiments.
For HAM10000, saturation occurs around 10 shots; however, to account for the sharper performance degradation observed below this threshold, we use 20 shots.
These choices result in assessment sets of 800 and 140 labeled samples per expert for CIFAR10-H and HAM10000, respectively.

In addition to $\mathcal{Y}_k$, our framework requires selecting two hyper-parameters: the optimal miscoverage rate $\alpha^*$ and the conformal scoring function (LAC, APS, or RAPS, with or without deterministic inclusion of the threshold-crossing class). These are selected using a held-out, expert annotated subset of the calibration data, referred to as the meta-validation set.

\begin{figure}[h]
    \centering
    \includegraphics[width=0.45\linewidth]{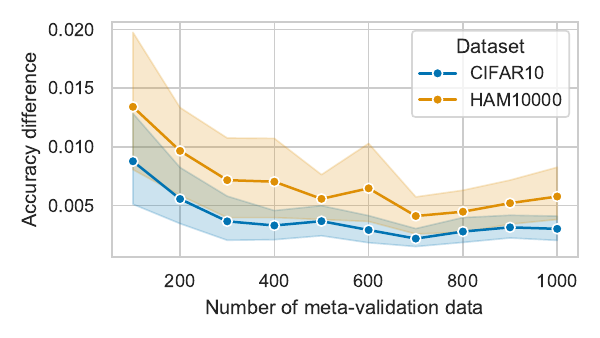}  
    \caption{Difference in system accuracy between optimal hyper-parameters selected in hindsight and those estimated using meta-validation sets of increasing size. Results are averaged over five random seeds and three model architectures; per-model trends are consistent. Shaded regions denote 95\% confidence intervals.}
    \label{fig:meta-val}
\end{figure}

The size of this meta-validation set introduces a trade-off between annotation cost and reliable hyper-parameter selection. Figure~\ref{fig:meta-val} shows the difference in system accuracy between optimal hyper-parameters selected in hindsight and those estimated using meta-validation sets of increasing size. As expected, this difference decreases as the meta-validation set grows and plateaus at approximately 700 samples for both datasets. Based on this saturation behavior, we use 700 samples as the meta-validation set size in all experiments.

\end{document}